%% file: Semantic-Aware_Generative_Adversarial_Nets_for_Unsupervised_Domain_Adaptation_in_Chest_X-ray_Segmentation.tex
\pdfoutput=1
\documentclass{llncs}

\usepackage{amsmath,amssymb,bm}
\usepackage{booktabs}

\usepackage{bbm}
\usepackage{gensymb}

\usepackage{url}
\usepackage{comment}
\usepackage{amsmath}
\usepackage{tabularx,threeparttable}
\usepackage{multirow}
\usepackage{graphicx}
\usepackage{bbm}
\usepackage{color}

\newcommand{\mD}{\mathcal{D}}
\newcommand{\mG}{\mathcal{G}}

\begin{document}

\title{Semantic-Aware Generative Adversarial Nets \\for Unsupervised Domain Adaptation \\in Chest X-ray Segmentation}
\vspace{-6mm}
\author{Cheng Chen\inst{1}, {Qi Dou\inst{1}}, Hao Chen\inst{1,2}, and Pheng-Ann Heng\inst{1}}

\institute{Dept. of Computer Science and Engineering, The Chinese University of Hong Kong
	\and Imsight Medical Technology Inc., China}

\maketitle              
\vspace{-4mm}
\label{sec:abs}
\input{abstract}

\vspace{-6mm}
\section{Introduction}
\vspace{-2mm}
\label{sec:intro}
\input{introduction}

\vspace{-1mm}
\section{Method}
\vspace{-2mm}
\label{sec:method}
\input{method}

\vspace{-1mm}
\section{Experimental Results}
\vspace{-0.5mm}
\label{sec:exp}
\input{experiments}

\vspace{-2mm}
\section{Conclusion}
\label{sec:conclusion}
\input{conclusion}

\vspace{-2mm}
\bibliographystyle{splncs03}
\bibliography{reference}
\vspace{-5mm}

\end{document}

%% file: abstract.tex
\begin{abstract}
In spite of the compelling achievements that deep neural networks (DNNs) have made in medical image computing, these deep models often suffer from degraded performance when being applied to new test datasets with domain shift.
In this paper, we present a novel unsupervised domain adaptation approach for segmentation tasks by designing semantic-aware generative adversarial networks (GANs).
Specifically, we transform the test image into the appearance of source domain, with the semantic structural information being well preserved, which is achieved by imposing a nested adversarial learning in semantic label space.
In this way, the segmentation DNN learned from the source domain is able to be directly generalized to the transformed test image, eliminating the need of training a new model for every new target dataset. 
Our domain adaptation procedure is unsupervised, without using any target domain labels.
The adversarial learning of our network is guided by a GAN loss for mapping data distributions, a cycle-consistency loss for retaining pixel-level content, and a semantic-aware loss for enhancing structural information.
We validated our method on two different chest X-ray public datasets for left/right lung segmentation.
Experimental results show that the segmentation performance of our unsupervised approach is highly competitive with the upper bound of supervised transfer learning.
\end{abstract}

%% file: introduction.tex
Deep neural networks (DNNs) have achieved great success in automated medical image segmentation, attributing to their learned highly-representative features.
However, due to domain shift, DNNs would suffer from performance degradation when being applied to new datasets, which are acquired with different protocols or collected from different clinical centers~\cite{ghafoorian2017transfer,kamnitsas2017unsupervised}. 
Actually, domain adaptation has been an important research topic in medical image computing and the traditional automated methods also encountered the same poor generalization problem.
For example, Philipsen \emph{et al.}~\cite{philipsen2015localized} studied the influence of data distribution variations across chest radiography datasets on traditional segmentation methods based on k-nearest neighbor classification as well as active shape modeling.

To generalize DNNs trained on a \emph{source domain} to a \emph{target domain}, researches have been emerging for domain adaptation of deep learning models.
A typical method is supervised transfer learning (STL), which fine-tunes the pre-trained source domain model with additional labeled target domain data.
Remarkably, Ghafoorian \emph{et al.}~\cite{ghafoorian2017transfer} studied on the number of fine-tuned layers to reduce the required amount of annotations for brain lesion segmentation across MRI datasets. 
However, the STL approaches still rely on extra labeled data, which is expensive or sometimes infeasible to obtain.

Instead, unsupervised domain adaptation (UDA) is more appealing to generalize models in clinical practice.
Early works have employed histogram matching to make test data resemble the intensity distribution of source domain data~\cite{wang1998correction}.
Recently, the generative adversarial networks (GANs) have made great achievements in generating realistic images and adversarial learning excels in mapping data distributions for domain adaptation~\cite{goodfellow2014generative,bousmalis2017unsupervised,ganin2016domain,tzeng2017adversarial}. 
In medical field, adversarial frameworks have been proposed to align feature embeddings between source and target data and presented promising results on cross-protocol brain lesion segmentation~\cite{kamnitsas2017unsupervised} and cross-modality cardiac segmentation~\cite{dou2018unsupervised}.
Recent works adopted CycleGAN~\cite{DBLP:conf/iccv/ZhuPIE17} as a data augmentation step to synthesize images from source domain to target domain, and used the pair of synthetic image and corresponding source label to train a segmentation model for target domain~\cite{chartsias2017adversarial,huo2017adversarial}. 
However, the synthetic images can be distorted on semantic structures, because the pure CycleGAN did not explicitly constrain the output of each single generator inside the cycle.
In addition, these methods require training a new model for every new target domain, which will bring burden to practical applications.

In this work, we propose a semantic-aware generative adversarial networks for unsupervised domain adaptation (named \emph{SeUDA}) of medical image segmentation.
Our method detaches the segmentation DNN from the domain adaptation process, and does not require any label from the test set.
Given a test image, our \emph{SeUDA} framework conducts image-to-image transformation to generate a source-like image which is directly forwarded to the established source DNN.
To enhance the preservation of structural information during image transformation, we improve CycleGAN with a novel semantic-aware loss by embedding a nested adversarial learning in semantic label space.
We validated our \emph{SeUDA} on two different chest X-ray public datasets for lung segmentation.
The performance of our unsupervised method exceeds the UDA baseline and is highly competitive with that of the supervised transfer learning.
Last but not least, our transformed image results are visually observable, which sheds the light on the explicit intuition of our proposed method. 

%% file: method.tex
Given the source domain images $x^s \! \in \! \mathcal{X}^s$ and the corresponding labels $y^s \! \in \! \mathcal{Y}$, we train a DNN model, denoted by $f^s$, which learns to segment the input images.
In UDA, we have unlabeled target images $x^t \! \in \! \mathcal{X}^t$ 
whose intensity distributions (or visual appearances) are not the same as the source domain data.
Fig.~1 presents an overview of our proposed \textit{SeUDA} framework, which adapts the appearance of $x^t$ to source image space $\mathcal{X}^s$, so that the established $f^s$ can be directly generalized to the transformed image.

\vspace{-2mm}
\subsection{Segmentation Network Established on Source Domain}
\vspace{-2mm}
Our segmentation DNN model (referred as segmenter) is detached from the learning of our domain adaptation GANs.
Compared with the integrated approaches, this advantage of an independent segmenter enables much more flexibility when designing a high-performance network architecture.
In this regard, we establish a state-of-the-art segmentation network which makes complementary use of the residual connection, dilated convolution and multi-scale feature learning~\cite{ronneberger2015u}.

The backbone of our segmenter is modified ResNet-101. We replace the standard convolutional layers in the high-level residual blocks with the dilated convolutions.
To leverage features with multi-scale receptive fields, we replace the last fully-connected layer with four parallel $3 \! \times \! 3$ dilated convolutional branches, with a dilation rate of \{6, 12, 18, 24\}, respectively.
An upsampling layer is added in the end to produce dense predictions for the segmentation task.
We start with 32 feature maps in the first layer and double the number of feature maps when the spatial size is halved or the dilation convolutions are utilized. The segmenter is optimized by minimizing the pixel-wise multi-class cross-entropy loss of the prediction $f^s(x^s)$ and ground truth $y^s$ with standard 
stochastic gradient descent.

\begin{figure*}[t]
	\label{fig:cmp}
	\includegraphics[width=\textwidth]{{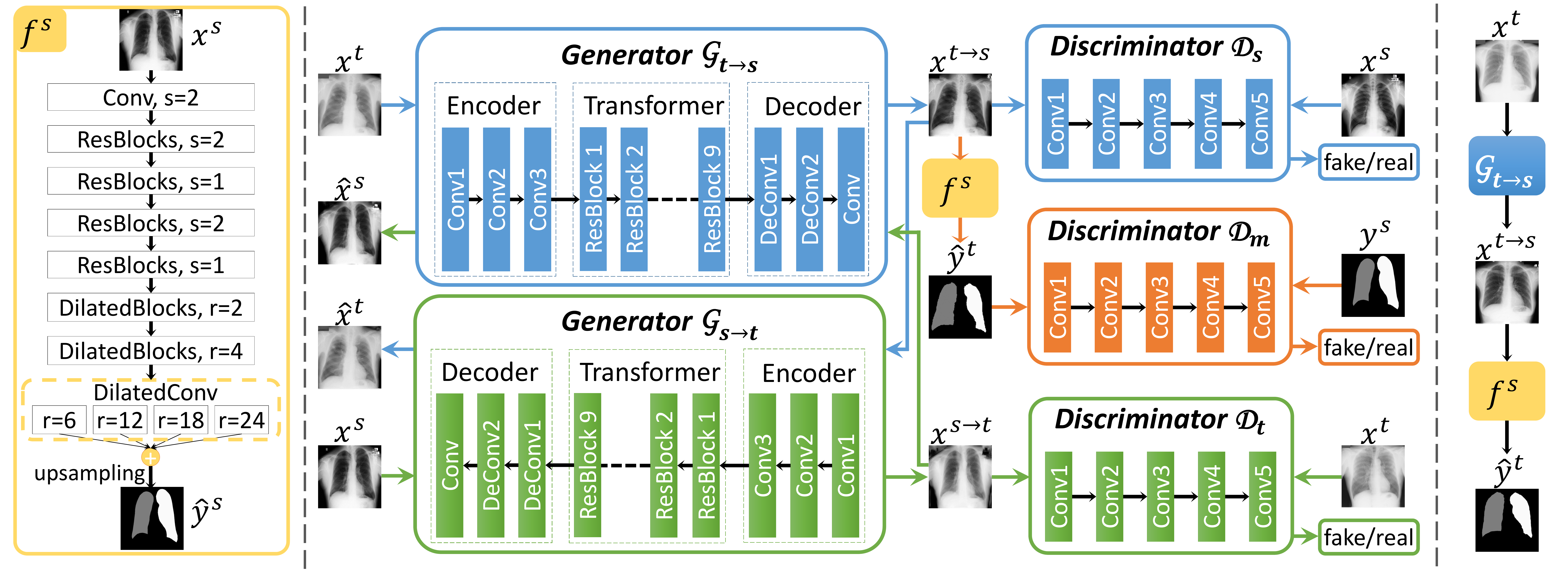}}
	\vspace{-6mm}
	\caption{The overview of our unsupervised domain adaptation framework. \textbf{Left}: the segmentation DNN learned on source domain; \textbf{Middle}: the \emph{SeUDA} where the paired generator and discriminator are indicated with the same color, the blue and green arrows respectively illustrate the data flows of $x^t$ and $x^s$ in cycle-consistency loss, the orange component is the discriminator for the semantic-aware adversarial learning;
    \textbf{Right}: the inference process of \emph{SeUDA} given a new target image during testing.
	}
    \vspace{-3mm}
\end{figure*}

\vspace{-3mm}
\subsection{Image Transformation with Semantic-Aware GANs}
\vspace{-2mm}
After obtaining $f^s$ which maps the source input space $\mathcal{X}^s$ to the semantic label space $\mathcal{Y}$, our goal is to make it generally applicable to new target datasets.
Given that annotating medical data is quite expensive, we conduct the domain adaptation in an unsupervised manner.
Specifically, we map the target images towards the source data space. The generated new image $x^{t\to s}$ appears to be drawn from $\mathcal{X}^s$ while the content and semantic structures remain unchanged.
In this way, we can directly apply the well-established model $f^s$ on $x^{t \to s}$ without re-training and get the segmentation result for $x^t$.

To achieve this, 
we first construct a generator $\mG_{t \to s}$ and a discriminator $\mathcal{D}_s$. The generator aims to produce realistic transformed image $x^{t\to s} = \mG_{t \to s}(x^t)$.
The discriminator competes with the generator by trying to distinguish between the fake generated data $x^{t \to s}$ and the real source data $x^s$.
The GAN corresponds to a minimax two-player game and is optimized via the following objective:
\begin{equation}
\vspace{-1mm}
\small
\mathcal{L}_{\text{GAN}}(\mG_{t \to s}, \mD_s)=\mathbb{E}_{x^s}[\text{log} \mD_s(x^s)]+\mathbb{E}_{x^t}[\text{log}(1-\mD_s(\mG_{t \to s}(x^t)))],
\label{eq:ganloss}
\end{equation}
where the discriminator tries to maximize this objective to correctly classify the $x^{t\to s}$ and $x^s$, while the generator tries to minimize $\text{log}(1-\mD_s(\mG_{t \to s}(x^t)))$ to learn the data distribution mapping from $\mathcal{X}^t$ to $\mathcal{X}^s$. 
\vspace{-1mm}
\\
\\
\textbf{Cycle-consistency adversarial learning.}
For image transformation,
the generated $x^{t\to s}$ should also preserve the detailed contents in the original $x^t$.
Inspired by the CycleGAN~\cite{DBLP:conf/iccv/ZhuPIE17} which sets the state-of-the-art for unpaired image-to-image transformation, we employ the cycle-consistency loss during the adversarial learning to maintain the contents with clinical clues of the target images.

Inversely, we build a source-to-target generator $\mG_{s \to t}$ and a target discriminator $\mD_t$, so that the transformed image can be translated back to the original image.
This pair of models are trained with a same-way GAN loss $\mathcal{L}_\text{GAN}(\mG_{s\to t}, \mD_t)$ following the Eq.~(\ref{eq:ganloss}).
In this regard, we derive the cycle-consistency loss which encourages $\mG_{s \to t}(\mG_{t \to s}(x^t)) \! \approx  \! x^t$ and $\mG_{t \to s}(\mG_{s \to t}(x^s)) \! \approx \! x^s$ in the transformation:
\begin{equation}
\vspace{-2mm}
\small
\begin{split}
\mathcal{L}_{\text{cyc}}(\mG_{t \to s}, \mG_{s \to t})= \mathbb{E}_{x^t}[||\mG_{s \to t}(\mG_{t \to s}(x^t))-x^t||_1]\! + \! \mathbb{E}_{x^s}[||\mG_{t \to s}(\mG_{s \to t}(x^s))-x^s||_1],
\vspace{-2mm}
\end{split}
\end{equation}
where the L1-Norm is employed for reducing blurs in the generated images. 
This loss imposes the pixel-level penalty on the distance between the cyclic transformation result and the input image.
\vspace{-1mm}
\\
\\
\textbf{Semantic-aware adversarial learning.}
In our proposed \emph{SeUDA}, we apply the established $f^s$ to $x^{t \to s}$ which is obtained by inputting $x^{t}$ to $\mG_{t\to s}$.
The image quality of $x^{t \to s}$ and the stability of $\mG_{t \to s}$ are crucial for the effectiveness of our method.
Therefore, besides the cycle-consistency loss which composes both generators and constraints the cyclic input-output consistency, we further try to explicitly enhance the intermediate transformation result $x^{t \to s}$.
Specifically, for our segmentation domain adaptation task, we design a novel semantic-aware loss which aims to prevent the semantic distortion during the image transformation.

In our unsupervised learning scenario, we establish a nested adversarial learning module by adding another new discriminator $\mD_{m}$ into the system.
It distinguishes between the source domain ground truth lung mask $y^s$ and the predicted lung mask $f^s(x^{t\to s})$ obtained by applying the segmenter on the source-like transformed image.
Our underlying hypothesis is that the shape of anatomical structure is consistent across multi-center medical images. 
The prediction of $f^s(x^{t\to s})$ should follow the regular semantic structures of the lung to fool the $\mD_{m}$, otherwise, the generator $\mG_{t\to s}$ would be penalized by the semantic-aware loss:
\begin{equation}
\vspace{-1mm}
\small
\mathcal{L}_{\text{sem}}(\mG_{t \to s}, \mD_m)= \mathbb{E}_{y^s}[\log\mD_m(y^s)]+\mathbb{E}_{x^t}[\log(1-\mD_m(f^s(\mG_{t \to s}(x^t))))].
\vspace{-1mm}
\end{equation}
This loss imposes an explicit constraint on the intermediate result of the cyclic transformation.
Its gradients can assist the update of the generator $\mG_{t\to s}$, which benefits the stability of the entire adversarial learning procedure.

\vspace{-2mm}
\subsection{Implementations and Training Procedure of SeUDA}
\vspace{-2mm}
The configurations of the generators and discriminators follow the practice of \cite{DBLP:conf/iccv/ZhuPIE17} and are presented in Fig.~1.
Specifically, both generators have the same architecture consisting of an encoder, a transformer and a decoder. 
All the three discriminators process $70 \! \times \! 70$ patches and produce real/fake predictions via 5 convolutional layers.
In our \emph{SeUDA}, the generators $\{\mG_{t\to s}, \mG_{s\to t}\}$ and the discriminators $\{\mD_s,\mD_t,\mD_m\}$ are optimized altogether and updated successively. Note that we do not update the segmenter $f^s$ in the process of image transformation.
The overall objective is as follows:
\begin{equation}
\small
\vspace{-1mm}
\begin{split}
\mathcal{L}(\mG_{s \to t}, \mG_{t \to s}, \mD_s, \mD_t, \mD_m)=
&~~\mathcal{L}_{GAN}(\mG_{s \to t}, \mD_t)+\alpha\mathcal{L}_{GAN}(\mG_{t \to s}, \mD_s)~+\\
&~\beta\mathcal{L}_{\text{cyc}}(\mG_{t \to s}, \mG_{s \to t})+\lambda\mathcal{L}_{\text{sem}}(\mG_{t \to s}, \mD_m),
\end{split}
\label{eq:totalloss}
\vspace{-1mm}
\end{equation}
where the $\{\alpha, \beta, \lambda\}$ denote trade-off hyper-parameters adjusting the importance of each component, which was set to be $\{0.5, 10, 0.5\}$ in our experiments.
The entire framework is optimized to obtain:
\begin{equation}
\small
\vspace{-1mm}
\begin{split}
\mG_{s \to t}^{\ast}, \mG_{t \to s}^{\ast} = \arg\min\limits_{\substack{\mG_{s \to t} \\ \mG_{t \to s}}}\max\limits_{\substack{\mD_s,\mD_t,\mD_m}}\mathcal{L}(&\mG_{s \to t}, \mG_{t \to s}, \mD_s, \mD_t, \mD_m).
\end{split}
\label{eq:optobj}
\vspace{-1mm}
\end{equation}

In practice, when training the generative adversarial networks,
we followed the strategies of~\cite{DBLP:conf/iccv/ZhuPIE17} for reducing model oscillation. Specifically, the negative log likelihood in $\mathcal{L}_{GAN}$ was replaced by a least-square loss to stabilize the training. 
The discriminator loss was calculated using one image from a collection of fifty previously generated images rather than the one produced in the latest training step.
We used the Adam optimizer with an initial learning rate of 0.002, which was linearly decayed every 100 epochs. We implemented our proposed framework on the TensorFlow platform using an Nvidia Titan Xp GPU.

%% file: experiments.tex
\vspace{-1.5mm}
\textbf{Datasets and Evaluation Metrics.}
We validated our unsupervised domain adaptation method for lung segmentations using two public Chest X-ray datasets, i.e., the Montgomery set (138 cases) \cite{jaeger2014two} and the JSRT set (247 cases)~\cite{shiraishi2000development}.
Both the datasets are typical X-ray scans collected in clinical practice, but their
image distributions are quite different in terms of the disease type, intensity, and contrast (see the first and fourth column in Fig.~2(a)). 
The ground truth masks of left and right lungs are provided in both datasets. 
We randomly split each dataset into 7:1:2 for training, validation and testing sets. All the images were resized to $512 \! \times \! 512$, and rescaled to $[0,255]$. The prediction masks were post-processed with the largest connected-component selection and hole filling.

For evaluation metrics, we utilized four common segmentation measurements, i.e.,
the Dice coefficient ([\%]), recall ([\%]), precision ([\%]) and average surface distance (ASD)([mm]). 
The first three metrics are measured based on the pixel-wise classification accuracy.
The ASD assesses the model performance at boundaries and a lower value indicates better segmentation performance.
\\
\\
\textbf{Experimental Settings.}
We employed the Montgomery set as source domain and the JSRT set as target domain.
We first established the segmenter on source training data independently.
Next, we test the segmenter under various settings:
1) testing on source domain (\textit{S-test}); 2) directly testing on target data (\textit{T-noDA}); 
3) using histogram matching to adjust target images before testing (\textit{T-HistM});
4) aligning target features with the source domain as proposed in~\cite{kamnitsas2017unsupervised} (\emph{T-FeatDA});
5) fine-tuning the model on labeled target data before testing on JSRT (\textit{T-STL});
In addition, we investigated the performance of our proposed domain adaptation method with and w/o the semantic-aware loss, i.e., \emph{SeUDA} and \emph{CyUDA}.
\vspace{-0mm}
\\
\vspace{-2mm}
\\
\begin{table*}[t]
	\centering
	\label{tab:results}
	\caption{Quantitative evaluation results of domain adaptation methods for both lung segmentations from chest X-ray images.}
	\vspace{-6mm}
	\renewcommand{\arraystretch}{1.1}
	\begin{center}
		\begin{tabular}{ |m{2cm}|m{1.1cm}|m{1.15cm}|m{1.25cm}|m{1.1cm}|m{1.1cm}|m{1.15cm}|m{1.25cm}|m{1.1cm}|  }
			\hline
			~\multirow{2}{*}{Methods} &\multicolumn{4}{c|}{Right Lung} & \multicolumn{4}{c|}{Left Lung}\\
			\cline{2-9}
			&~~Dice&~Recall&Precision&~~ASD &~~Dice &~Recall& Precision&~~ASD\\
			\hline
			~S-test &~~95.98&~~97.98&~~94.23&~~2.23& ~~95.23 & ~~96.56& ~~94.01 &~~2.45\\
			\hline
			~T-noDA &~~82.29&~~98.40&~~73.38&~~10.68& ~~76.65 &~~95.06&~~69.15&~~11.40\\
			~T-HistM \cite{wang1998correction} & ~~90.05 &~~92.96&~~88.05&~~5.72& ~~91.03 &~~94.35&~~88.45&~~4.66\\
			~T-FeatDA\cite{kamnitsas2017unsupervised} & ~~94.85 &~~93.66&~~96.42&~~3.26&~~92.93&~~91.67&~~94.46&~~3.80\\
			~T-STL \cite{ghafoorian2017transfer} &~~96.91&~~98.47&~~95.46&~~1.93& ~~95.84 &~~97.48&~~94.29&~~2.20\\
			\hline
			~CyUDA & ~~94.09 & ~~96.31& ~~92.28  & ~~3.88& ~~91.59 & ~~92.28& ~~91.70  & ~~4.57 \\
			~SeUDA & ~~{95.59}  & ~~{96.55}& ~~{94.77} & ~~{2.85} & ~~{93.42} & ~~{92.40}& ~~{94.70}& ~~{3.51} \\
			\hline
		\end{tabular}
	\end{center}
	\vspace{-7mm}
\end{table*}
\\
\vspace{-8mm}
\\
\textbf{Comparison of Experimental Results Between Different Methods.}
As shown in Table~1, when directly applying the learned source domain segmenter to target data~(\emph{T-noDA}),
the model performance significantly degraded, indicating that domain shift would severely impede the generalization performance of DNNs.
Specifically, the average Dice over both lungs dropped from 95.61\% to 79.47\%, and the average ASD increased from 2.34 to 11.04~$mm$.

With our proposed \emph{SeUDA}, remarkable improvements have been achieved by applying the source segmenter on transformed target images.
Compared with \emph{T-noDA}, our \textit{SeUDA} increased the average Dice by 15.04\%.
Meanwhile, the ASDs for both lungs were reduced significantly.
Also, our method outperforms the UDA baseline histogram matching \emph{T-HistM} with the average dice increased by 3.97\% and average ASD decreased from 5.19~$mm$ to 3.18~$mm$. Compared with the feature-level domain adaptation method \emph{T-FeatDA}, our \emph{SeUDA} can not only obtain higher segmentation performance, but also provide intuitive visualization of how the adaptation is achieved.
Notably, the performance of our unsupervised \textit{SeUDA} is even comparable to the upper bound of supervised \textit{T-STL}.
In Table~1, the gaps of Dice are marginal, i.e., 1.32\% for right lung and 2.42\% for left lung.

In Fig.~2(a), we can visualize typical transformed target images, demonstrating that \textit{SeUDA} has successfully adapted the appearance of target data to look more similar to source images. In addition, the positions, contents, semantic structures and clinical clues are well preserved after transformation. In Fig.~2(b) we can observe that without domain adaptation, the predicted lung masks are quite cluttered. With histogram matching, appreciable improvements are obtained but the transformed images cannot mimic the source images very well. With our \textit{SeUDA}, the lung areas are accurately segmented attributing to the successful target-to-source appearance transformation.
\begin{figure*}[t]
	\label{fig:cmp}
	\includegraphics[width=1.0\textwidth]{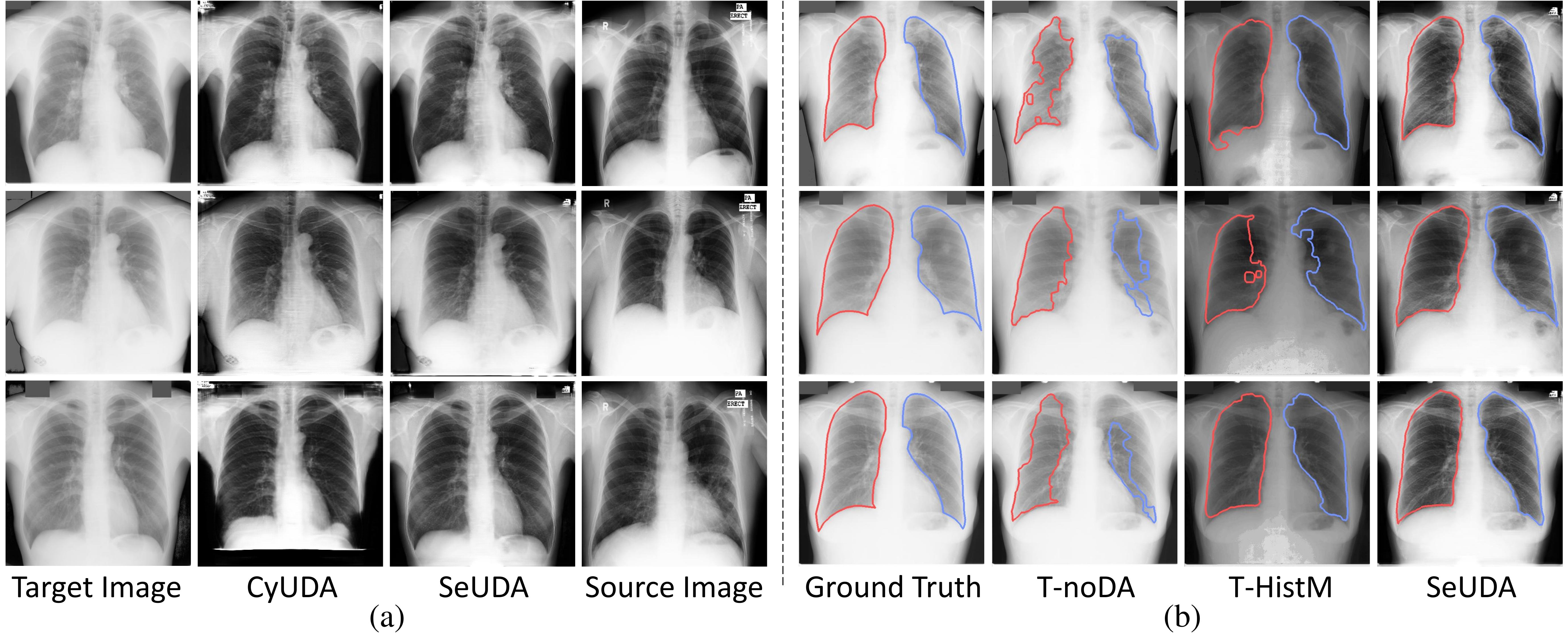}
	\vspace{-6mm}
	\caption{Typical results for the image transformation and lung segmentation. (a) Visualization of image transformation results, from left to right, are the target images in JSRT set, \textit{CyUDA} transformation results, \textit{SeUDA} transformation results, and the nearest neighbor of $x^{t\to s}$ got from source set; each row corresponds to one patient. (b) Comparison of segmentation results between the ground truth, \textit{T-noDA}, \textit{T-HistM}, and our proposed \textit{SeUDA}; each row corresponds to one patient.} 
	\vspace{-4mm}
\end{figure*}
\\
\vspace{-2mm}
\\
\textbf{Effectiveness of Semantic-aware Loss.}
We investigated the contribution of our novel semantic-aware loss designed for segmentation domain adaptation.
We implemented \textit{CyUDA} by removing the semantic-aware loss from the \emph{SeUDA}.
One notorious problem of GANs is that their training would be unstable and sensitive to initialization states~\cite{bousmalis2017unsupervised,salimans2016improved}. 
In this study, we measured the standard deviation (std) of the \textit{CyUDA} and \textit{SeUDA} by running each model for 10 times under different initializations but with the same hyper-parameters.
We observed significant lower variability on the segmentation performance across the 10 \textit{SeUDA} models than the 10 \textit{CyUDA} models, i.e., Dice std: 0.25\% v.s. 2.03\%, ASD std: 0.16 v.s. 1.19~$mm$. 
Qualitatively, we observe that the \textit{CyUDA} transformed images may suffer from distorted lung boundaries in some cases, see the third row in Fig.~2(a).
In contrast, adding the semantic-aware loss, the transformed images consistently present a high quality.
This reveals that the novel semantic-aware loss contributes to stabilize the image transformation process and prevent the distortion in structural contents, and hence contributes to boost the performance of segmentation domain adaptation.

%% file: conclusion.tex
\vspace{-2mm}
This paper proposes a novel approach \emph{SeUDA} for unsupervised domain adaptation of medical image segmentation.
The \textit{SeUDA} leverages GANs to transform the target images to resemble source images and generalize the source segmentation DNN directly on the transformed images. 
We design a novel objective which composes a GAN loss for mapping data distributions, a cycle-consistency loss to preserve the pixel-level content, and a semantic-aware loss to enhance the structural information.
Our method is highly competitive with the supervised transfer learning on the task of lung segmentation in chest X-rays.
Our proposed SeUDA solution is general and can inspire more researches on domain adaptation problems in medical image computing.